\documentclass[journal]{IEEEtran}
\usepackage{amsmath,amsfonts}
\usepackage{algorithm}
\usepackage{array}
\usepackage[caption=false,font=normalsize,labelfont=sf,textfont=sf]{subfig}
\usepackage{textcomp}
\usepackage{stfloats}
\usepackage{url}
\usepackage{verbatim}
\usepackage{graphicx}
\usepackage{cite}
\usepackage{multirow}
\usepackage{makecell}
\usepackage{amssymb}
\usepackage{amsmath}
\usepackage{color}
\usepackage{url}
\usepackage{setspace}
\usepackage{threeparttable}
\usepackage{algpseudocode}
\usepackage{mathrsfs}
\usepackage{graphicx,graphics}
\usepackage{lipsum,epsfig,dsfont}
\begin{document}

\title{EiHi Net: Out-of-Distribution Generalization Paradigm}

\author{Qinglai~Wei,~\IEEEmembership{Senior Member,~IEEE,}
        Beiming~Yuan
        and~Diancheng~Chen

\thanks{This work was supported in part by the National Key R\&D Program of China under Grants 2021YFE0206100;
  in part by the National Natural Science Foundation of China under Grant 62073321;
  in part by National Defense Basic Scientific Research Program JCKY2019203C029;
  in part by the Science and Technology Development Fund, Macau SAR under Grant
  0015/2020/AMJ. (Corresponding author: Diancheng Chen)
}

\thanks{Q. Wei is with the State Key Laboratory for Management and Control of Complex Systems, Institute of Automation, Chinese Academy of Sciences, Beijing 100190, and with the School of Artificial Intelligence, University of Chinese Academy of Sciences, Beijing 100049, and also with the Institute of Systems Engineering, Macau University of Science and Technology, Macau 999078, China (e-mail: qinglai.wei@ia.ac.cn).}

\thanks{B. Yuan is with the School of Artificial Intelligence, University of Chinese Academy of Sciences, Beijing 100049, China (e-mail: yuanbeiming20@mails.ucas.ac.cn).}

\thanks{D. Chen is with the State Key Laboratory for Management and Control of Complex Systems, Institute of Automation, Chinese Academy of Sciences, Beijing 100190, and also with the University of Chinese Academy of Sciences, Beijing 100049, China (e-mail: chendiancheng2020@ia.ac.cn).}

\thanks{Q. Wei and B. Yuan contributed equally to this work.}

}

\markboth{Submitted to IEEE Transactions on Neural Networks and Learning Systems}%
{Shell \MakeLowercase{\textit{et al.}}: A Sample Article Using IEEEtran.cls for IEEE Journals}


\maketitle

\begin{abstract}
This paper develops EiHi to solve the out of distribution (O.o.D.) problem in recognition, especially the distribution shift of background. EiHi net is a model learning paradigm that can be blessed on any visual backbone. Different from the traditional deep model, namely find out correlated relationship between sample features and corresponding categories, EiHi net suffers less from spurious correlations. Fusing SimCLR and VIC-Reg via explicit and dynamical establishment of original - positive - negative sample pair as a minimal learning element, EiHi net iteratively establishes a relationship close to the causal one between features and labels, while suppresses the spurious correlations. To further reinforce the ability of the proposed model on causal reasoning, we develop a human-in-the-loop strategy, with few guidance samples, to prune the representation space directly. Finally, it is shown that the developed EiHi net makes significant improvements in the most difficult and typical O.o.D. dataset NICO, compared with the current SOTA results, without any domain ($e.g.$ background) information.
\end{abstract}


\section{Introduction}
\IEEEPARstart{T}{he} performance of traditional deep learning algorithms \cite{ref1,ref2} is excellent when the image samples of training and test sets are independent and identically distributed (I.I.D.) \cite{ref3,ref4,ref5,ref6}. However, the assumption of I.I.D is fragile and easy to break. When it disappears, deep models show a decline in performance in several tasks \cite{ref7,ref8}. The O.o.D. problem in image learning is that the collected data are semantically grouped by human labels, but the distributions in pixel-level in each group is quite different. In particular, the background information of image data is a common form of O.o.D.

Tracking down the reason of decline, we notice that traditional deep learning algorithms are based on inducing correlations between features and categories. When the background of the data are biased, the knowledge summarized by the model on the training set is difficult to generalize \cite{ref9,ref10}. However, the background O.o.D problem has little impact on humans, and we believe that it is closely related to the way humans learn. When determining the category of objects, humans are capable of filtering irrelevant background information and looking for unique features of the target. In view of this situation, we design two paradigms to solve the background O.o.D. problem, by mimicking human learning style, and starting from two key O.o.D. factors, namely {Diversity Shift} and {Correlation Shift} \cite{ref11}. Diversity Shift refers to the offset between the backgrounds of the training set and test set. Correlation shift refers to the offset of the background information in the training set. {Non-I.I.D. Image dataset with Contexts} (NICO) \cite{ref12} refers the background of image as `domain' and carries these two O.o.D. factors simultaneously. For instance, the trivial model is trained to identify the presence of sheep, but our training data do not have sheep standing in the snow. This is the case where {Diversity Shift} appears in the data. {Correlation Shift} \cite{ref18,ref17} will also bring fatal damage to the performance of the deep learning algorithm, such that the biased distribution usually causes the spurious correlation between the domain and the label \cite{ref11}. For example, most sheep data acquisition takes place on grasslands, and this situation will cause a spurious correlation between grassland and the sheep's own features, as most of the samples with "sheep" labels contain grassland and sheep simultaneously. To sum up, in traditional image classification tasks, the deep learning model trained by the cross-entropy loss function takes the correlation between sample features and classes as the basis of sample classification tasks \cite{ref8,ref20}, and cannot encode features of the target in the desired way under O.o.D. situation.

In the meantime, we noticed that children learn to distinguish various animals through mediums like cartoon cards and photos under proper guidance. This process gives us the inspiration for designing the EiHi (\textbf{E}liminate  \textbf{i}rrelevant features with \textbf{h}uman \textbf{i}ntervention) network. Note that EiHi contains two paradigms, and the first paradigm can be seen as application of supervised contrastive learning\cite{refworning} in O.o.D. settings, and the second paradigm is a human-in-the-loop strategy to deal with O.o.D. problems.

\section{Problem Description}

\begin{figure}[ht]\centering
	\includegraphics[width=8cm]{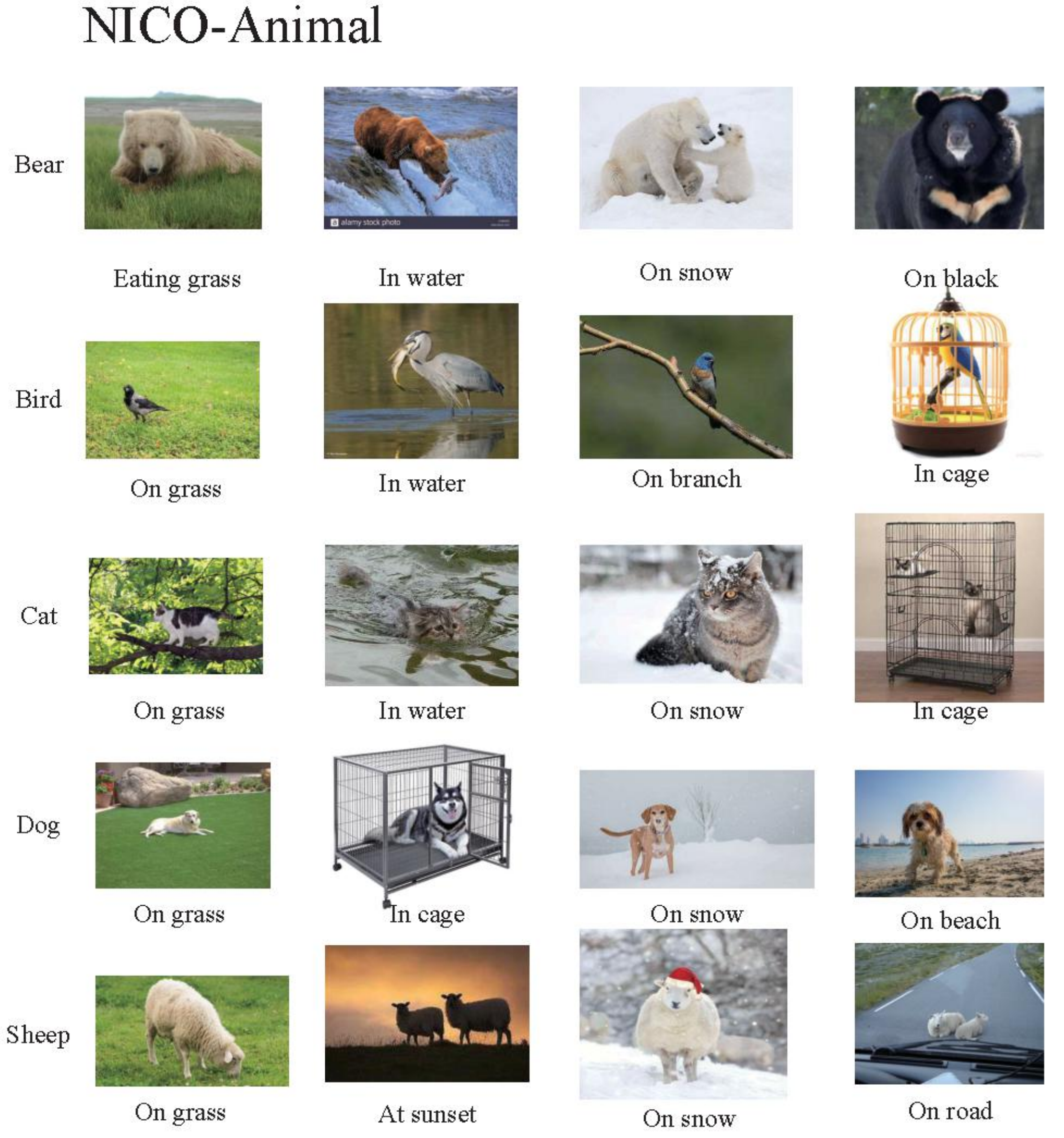}
	\caption{The figure shows the example of NICO-animal.}
\label{animal}
\end{figure}

\begin{figure}[ht]\centering
	\includegraphics[width=8cm]{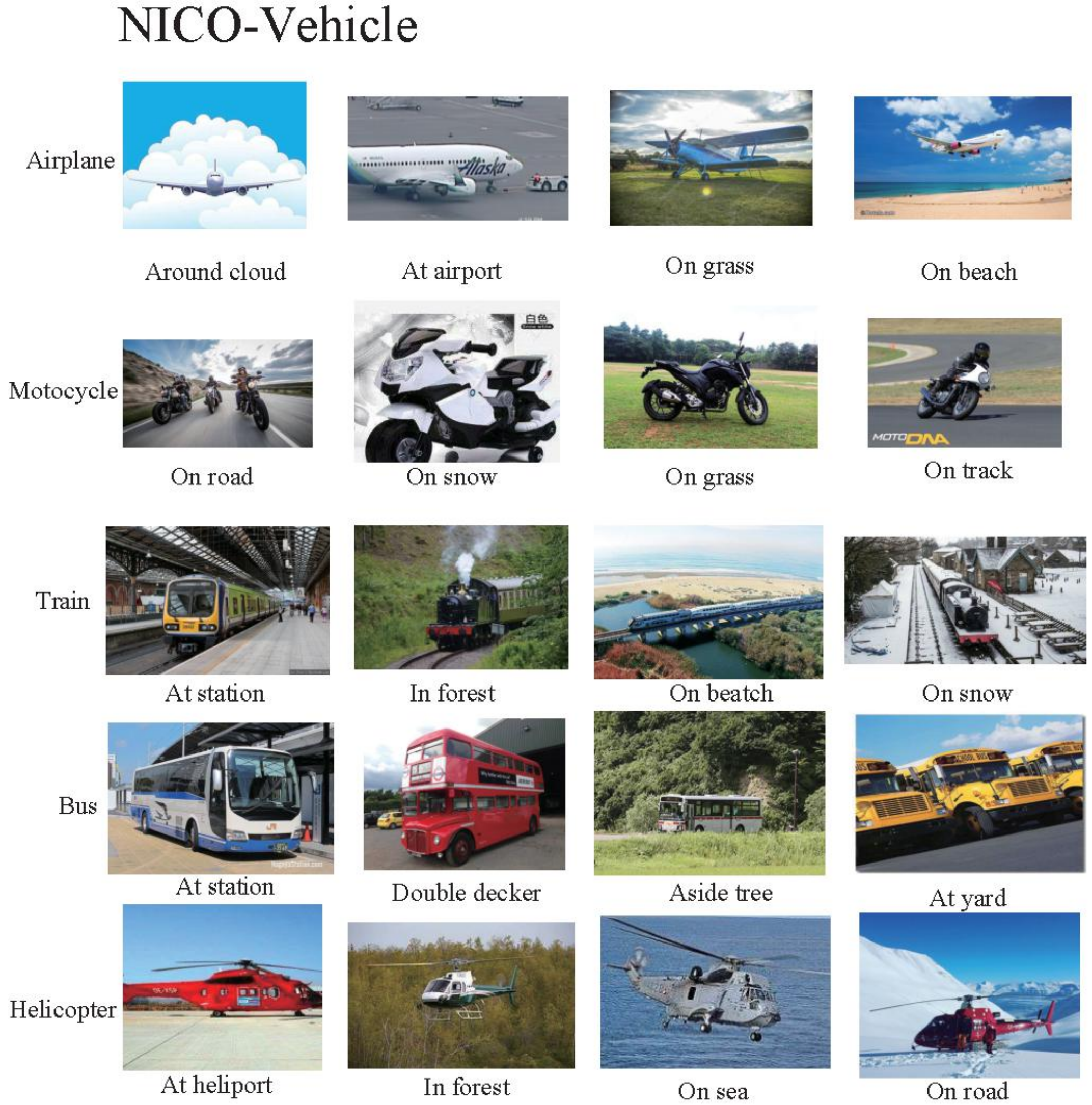}
	\caption{The figure shows the example of NICO-vehicle.}
\label{vehicle}
\end{figure}

We describe both the Diversity Shift and Correlation shift problems from the experimental perspective based on the NICO data set. There are 10 types of animals in the NICO-animal dataset, with each animal has 10 domains, including the background and pose of the animal. As for NICO-vehicle, there are 9 types of vehicles, each of which has 10 domains, including the scene and style. Some examples of the NICO-animal and NICO-vehicle are shown in Figures \ref{animal} and \ref{vehicle}.

\emph{Diversity Shift.} We randomly choose 8 domains as the source domains, and the remaining 2 domains as the target domains. We first train and test the deep learning model on all domains, and take the test accuracies as the benchmark. We then train the model with the source domain data and test it on the target domains, and compare the test results with the benchmark to observe the impact of Diversity Shift as the first experiment. To ensure the fairness of the experiment, we keep the settings of training sets and test sets consistent in all experimental setups. The experimental results shown in Table \ref{Impact_vehicle} indicates the huge impact of Diversity Shift on deep learning models.

\begin{table}[ht]
\centering
\caption{Experiment of Diversity Shift Impact (8:2)}
\begin{threeparttable}
\begin{tabular}{cccc}
\hline\hline
dataset &model         &\makecell[c]{Domain \\mixing(10:10)\tnote{1}} &\makecell[c]{Domain \\shift(8:2)\tnote{2}}\\
\hline

\multirow{4}*{NICO-Animal}&ResNet18  	   &41.03$\pm$0.10\% &28.56$\pm$0.30\%  \\

~&ResNet50	   &54.55$\pm$0.15\%  &39.55$\pm$0.25\%  \\

~&Vgg16       	&62.49$\pm$0.21\% &48.66$\pm$0.50\% \\ 

~&Vgg19       	&60.22$\pm$0.13\% &46.11$\pm$0.09\% \\ 

\hline

\multirow{4}*{NICO-Vehicle}&ResNet18  	   &53.45$\pm$0.12\% &42.29$\pm$0.08\%  \\

~&ResNet50	   &70.20$\pm$1.18\%  &55.16$\pm$0.35\% \\

~&Vgg16       	&74.13$\pm$1.33\% &58.24$\pm$0.40\%  \\

~&Vgg19       	&74.59$\pm$1.26\% &57.60$\pm$0.80\% \\

\hline\hline

\end{tabular}

\label{Impact_vehicle}
 \begin{tablenotes}
        \footnotesize
       \item[1]Separate training set and test set from mixed domians.
\item[2]The training set is from the source domain, and the test set is from the target domain.
 \end{tablenotes}

\end{threeparttable}

\end{table}

\emph{Correlation Shift.} In the experiment of investigating the influence of Correlation Shift \cite{ref18,ref17}, we magnify this phenomenon by randomly reducing samples in some training sets. Specifically, we select one of the 8 source domains as the primary domain and reduce the number of samples in the remaining 7 domains to {$1/5$} the amount of primary one as the secondary domain. The reduced version of source domains (primary domain with secondary domains) are used for training, while the target domain is left unchanged. The experimental results of vgg16 on NICO with stronger domain shift are shown in Table \ref{correlation shift impact}, and we make it as the benchmark for the Correlation Shift problem.

Because prevalent pre-training(e.g. ImageNet pre-training model) covers most of the information needed in our experiments, all experiments on NICO do not adopt the pre-training scheme. We conduct every experiment 5 times, and report the average of the 5 test accuracies.

We also tried some challenging problems which enhance the strength of both Diversity Shift and Correlation Shift to observe the performance of the developed EiHi net.

\begin{table}[ht]
\centering
\caption{Vgg16 on nico 8:2 with Correlation shift}
\begin{threeparttable}
\begin{tabular}{cccc}
\hline\hline
\makecell[c]{dataset}&model&\makecell[c]{original \\(8:2)\tnote{1}}&\makecell[c]{Stronger domain\\ shift(8:2)\tnote{2}} \\
\hline

\makecell[c]{NICO-Animal}&\makecell[c]{vgg16}	
&48.66$\pm$0.50\%&37.83$\pm$0.12\%\\

~&~	&~&~\\

\makecell[c]{NICO-vehicle}&\makecell[c]{vgg16}
&58.24$\pm$0.40\%  	&41.85$\pm$0.25\% \\ 

\hline\hline
\end{tabular}
\label{correlation shift impact}
 \begin{tablenotes}
        \footnotesize
       \item[1]Do not change the data volume proportion of the source domain.
\item[2]The data volume of the secondary domain in the source domain is reduced by $1/5$ of the primary domain.
 \end{tablenotes}

\end{threeparttable}
\end{table}

\section{Related works}
In this section, we will briefly introduce the research status of the O.o.D. problem.

\textbf{NICO dataset \cite{ref12}:} The problem of O.o.D. image classification has not been studied thoroughly before the emergence of well-designed datasets. NICO uses context to consciously create Non-Independent and Identically Distributed (Non-IID) data. Compared with other data sets, NICO can support various Non-IID situations with sufficient flexibility. A lot of works have contributed to eliminating the O.o.D. influence, $e.g.$ domain adaptation and covariate shift methods \cite{ref25,ref26,ref27}. however, these works are not feasible in practice, because they need the test data distribution as prior knowledge. In \cite{ref12}, multiple Non-I.I.D datasets are compared, and NICO is found to be the only dataset which can support different types of O.o.D, and boost researches on robust and explainable machine learning \cite{ref21,ref22,ref23,ref24}.

\textbf{Stable learning \cite{ref13}:} O.o.D problem states that models perform well on the training set do not necessarily perform well on test set. StableNet provides a novel method to solve such problems: by sample reweighting, the shift domain distribution in the samples is forcibly corrected, and the correlation between features is reduced. and the independence control of sample features is provided.

\textbf{Dec Aug \cite{ref11}:} Traditional methods either assume the known heterogeneity in training data (such as domain labels) or assume that the capacities of different domains are roughly equal. DecAug analyzes two kinds of O.o.D. factors in Nico data, describing Nico as a two-dimensional O.o.D. problem and expounding the necessity of adding a domain label. By introducing domain labels, DecAug arranges the deep learning model to extend two branches to learn domain information and object information in the sample, respectively. In addition, DecAug arranges the learning directions of the parameters of the two branches to be perpendicular, in order to ensure that the information decoupling. However, the introduction of domain labels increases the workload. In addition, many domains are difficult to categorize.

\textbf{SimCLR \cite{ref14}:} There are several key points in SimCLR: the combination of data augmentation; A learnable nonlinear projector is introduced to improve the quality of representation learning; The $infoNCE$ is adopted in this paper as loss function.

\textbf{VICReg \cite{ref16}:} The self-monitoring method for image representation learning is generally based on the consistency between the embedding vectors of the same image with different data augmentaions. The author introduces VICReg (Variance-Invariance-Covariance Regularization), which has three regularization terms on the embedded variance.

\textbf{Self-supervised Batch Comparison  \cite{refworning}:} In this work, the authors extend the self-supervised batch comparison method to the fully supervised setting so that the model can effectively use the label information.

\textbf{SCM \cite{refscm}:} SCM believe that when the Contrastive learning (CL) model is trained with the entire image, the performance is terrible in test set with only foreground region; when the CL model is trained with foreground region, the performance is also terrbile in test set with full image. This observation shows that the background in the image may interfere with the model learning, and their influence has not been completely eliminated. To solve this problem, the author builds a structural causal model (SCM) to model the background as a hybrid factor, and propose a regularization method based on back door adjustment to perform a causal intervention on the proposed SCM.

\section{Methdology}
\subsection{Feature Extraction}

Using fewer data, such as cards with animal pictures, people can learn to identify animals in any scene. Imagine a scene where parents hold animal painting cards, point to the cow on it, and describe "the one with horns is cow". In this scenario, We believe that there are two key points for effective learning with limited data. The first is to point out which features are unique to one class, and the second is the partition of objects, which removes spurious-relevant features \cite{ref8,ref20}.

In image classification tasks, traditional convolutional neural network (CNN) is usually composed of stacked convolutional layers and stacked fully connected neural networks, such as the ResNet \cite{ref2} series and Vgg \cite{ref21} series. The stacked convolution layer captures image features, in the following, we also call these layers feature capture networks. Features in the image will activate convolution kernels to produce representations, which will be used in the following stacked full connection layers to makes decisions. We design two paradigms to work on the above-mentioned process, so as to deal with O.o.D. problems. This subsection introduces the first paradigm, namely pairwise decorrelation contrastive learning.

To enable the model to find proper features belonging to each class, the core idea is to suppress the spurious causal relationships. We argue that, when the spurious causal feature (e.g., the background information) appears in both the positive and negative samples, it can serve as a weapon against O.o.D. problems, with increasing effectiveness as the amount of negative samples increases. In the process of pairwise comparison of original-positive-negative samples, the object information of the samples will be retained, and the domain information will be gradually abandoned, by pairing multiple negative samples to the positive counterpart.

A supervised version of VICReg to complete the original-positive-negative sample learning process is required for several reasons. VICReg \cite{ref16} sets up positive and negative samples explicitly and implicitly, respectively. However, this method is not suitable for solving the O.o.D. problem. The variance loss term implicitly defines the negative samples by implying that samples in each batch should be different, which is unfriendly to tiny dataset like NICO, where the possibility of different samples belonging to the same class in a batch is relatively high. Furthermore, VICReg prevents the dimensional collapse by covariance constraint. The covariance term expects a larger batch size, but such a size will introduce more positive samples in every batch, which brings forth contradictory since all of them are treated as negative samples in the perspective of the variance term. ADACLR\cite{ref22} also mentioned these issues. Inspired by \cite{refworning}, we abandon the implicit way of defining negative samples, reformulate VICReg into supervised manner, similar works can refer to\cite{refworning}. Specifically, we explicit set positive and negative samples based on the object label to train the feature capture network\cite{refworning}. Different samples of the same class are set as positive samples, and the samples from other classes naturally become negative counterparts.

Further refinement of VICReg is required to make it compatible with the new supervised setting. We propose to use ${\ell _{{\rm{infoNCE}}}}\left( \cdot \right)$ in Equation (\ref{loss}) to replace variance and invariance terms in VICReg, and keep covariance ${\ell _{{\rm{cov}}}}\left( \cdot \right)$ unchanged. Variance term is abandoned to avoid the contradiction aforementioned. As for invariance term, applying invariance term to encode different samples into identical features will induce another collapsed solution, which is remedied by variance term in original VICReg. In view of the abandonment of variance term, we further abandon the invariance term. Lastly, we keep the covariance term unchanged. Ablation studies are conducted to evaluate the effectiveness of the new loss combinations.

It is necessary to elucidate the difference between the proposed supervised VICReg and the SimCLR with its supervised version \cite{refworning}. In model structure level, first, our model in its current form collects original-positive-few negative samples as minimal learning element, while SimCLR and its supervised version collects each group of negative samples from each minibatch. Second, we do not adopt a projection head. In loss level, we combine ${\ell _{{\rm{infoNCE}}}}\left( \cdot \right)$ with covariance term, the former term acts on each minimal learning element, and the later term acts on the whole minibatch. While the loss terms in SimCLR with its supervised version act on the whole minibatch to develop positive and negative samples. In implementation level, it turns out that about 9 negative samples for each learning element will suffice, and we do not require large batchsize. In motivation level, we set negative samples explicitly to suppress the spurious correlations in O.o.D. problems, and we only introduce one positive sample for each original sample to avoid introducing more spurious correlations.

Here we define some notations. Original sample $ {{x_{or{i}}}} $, positive sample $ {{x_{pos}}} $ and $M$ negative samples $\left\{ {x_{ne{g^m}}} \right\}_{m = 1}^M$ (${x_{ori}},{x_{po{s}}},{x_{ne{g^m}}} \in {R^{c \times h \times w}}$) denoted as minimal learning element is feed to the backbone to get vector representation $ {{z_{or{i}}}} $, $ {{z_{pos}}} $ and $\left\{ {z_{ne{g^m}}} \right\}_{m = 1}^M$ (${z_{ori}},{z_{po{s}}},{z_{ne{g^m}}} \in {R^{d}}$), respectively. We use ${\ell _{{\rm{infoNCE}}}}\left( {{z_{ori}},{z_{pos}},{z_{ne{g^1}}},{z_{ne{g^2}}},...,{z_{ne{g^M}}}} \right)$ to draw $z_{ori}$ and $z_{pos}$ closer, and make directions of $z_{ori}$ and ${z_{ne{g^m}}}$ far away from each other, in terms of cosine similarity. And then, we use ${\ell _{{\rm{cov}}}}\left( \cdot \right)$ to reduce the correlation between each dimension of $z$ ($z = \left\{{z_{ori}},{z_{po{s}}},{z_{ne{g^m}}} \right\}$). The normalized feature representation with prevention of dimensional collapse effectively prevent trivial solutions in the network.

Our loss function setting is shown in the following equation:

\begin{align}\label{loss}
\ell\Big( {Z_{ori}},&{Z_{pos}},{Z_{ne{g^1}}},{Z_{ne{g^2}}},...,{Z_{ne{g^M}}} \Big) \nonumber \\
&=\lambda {\ell _{{\rm{info}}\_{\rm{nce}}}}\left( {{Z_{ori}},{Z_{pos}},{Z_{ne{g^1}}},{Z_{ne{g^2}}},...,{Z_{ne{g^M}}}} \right) \nonumber\\
&\quad\quad+{\ell _{{\rm{cov}}}}\left( {Z_{ori}} \right) +{\ell _{{\rm{cov}}}}\left( {{Z_{pos}}} \right)+{\ell _{{\rm{cov}}}}\left( {\textbf{Z}_{neg}} \right),
\end{align}
where ${\textbf{Z}_{neg}} =  \left\{{Z_{ne{g^m}}}\right\}_{m = 1}^M $, $ Z_{ori} = \left\{ {{z_{or{i_i}}}} \right\}_{i = 1}^n $,
$ Z_{pos} = \left\{ {{z_{po{s_i}}}} \right\}_{i = 1}^n $, ${Z_{ne{g^m}}}= \left\{ {z_{ne{g^m}_i}} \right\}_{i = 1}^n$,
${z_{ori}}_{_i},{z_{po{s_i}}},{z_{ne{g^m}_i}} \in {R^d}, \\m \in [1,M],  i \in [1,n]$, where $M$ is the amount of negative samples in each minimal learning element, and $n$ is the minibatch size. The function  ${\ell _{{\rm{infoNCE}}}}\big( {{Z_{ori}},{Z_{pos}},{Z_{ne{g^1}}},{Z_{ne{g^2}}},...,{Z_{ne{g^M}}}} \big)$ is expressed as
\begin{align} \label{loss-parameter}
{\ell _{{\rm{infoNCE}}}}\big( {{Z_{ori}},{Z_{pos}},{Z_{ne{g^1}}},{Z_{ne{g^2}}},...,{Z_{ne{g^M}}}} \big)  \\= - \frac{1}{n}\sum\limits_{i = 1}^n {\left( {\log \left( {si{m_i}} \right)} \right)},\nonumber
\end{align}
where
\begin{align}\label{nce}
&{si{m_i}}\nonumber\\&= softmax\left( {\frac{{si{m_{po{s_i}}}}}{t},\frac{{si{m_{ne{g^1}_i}}}}{t},\frac{{si{m_{ne{g^2}_i}}}}{t}...\frac{{si{m_{ne{g^M}_i}}}}{t}} \right).
\end{align}
The terms $si{m_{po{s_i}}}$ and $si{m_{ne{g^m}_i}}$, $m \in [1,M]$ are computed by ${si{m_{po{s_i}}}}= {||{Z_{or{i_i}}} \cdot {Z_{po{s_i}}}|{|_2}}$ and $si{m_{ne{g^m}_i}} = ||{Z_{or{i_i}}} \cdot {Z_{ne{g^m}_i}}|{|_2}$, respectively. The parameter $t \in {R}$ in Equation (\ref{nce}) is the temperature coefficient.

Next, we consider the function ${\ell _{{\rm{cov}}}}( {Z})$ in (\ref{loss}), which is
\begin{align}\label{cov}
&{\ell _{{\rm{cov}}}}({Z}) \nonumber\\ &=\frac{1}{d}\sum {\left( {{{\left( {\frac{1}{{n - 1}}\sum\limits_{i = 1}^{n} {\left( {{{\rm{Z}}_i} - \overline {\rm{Z}} } \right){{\left( {{{\rm{Z}}_i} - \overline {\rm{Z}} } \right)}^T}} } \right)}^2} \cdot \left( {1-I} \right)} \right)}
\end{align}
where
\begin{align}\label{z_avg}
\overline Z  = \frac{1}{n}\sum\limits_{i = 1}^{n} {{Z_i}}.
\end{align}

\begin{figure}[ht]\centering
	\includegraphics[width=8cm]{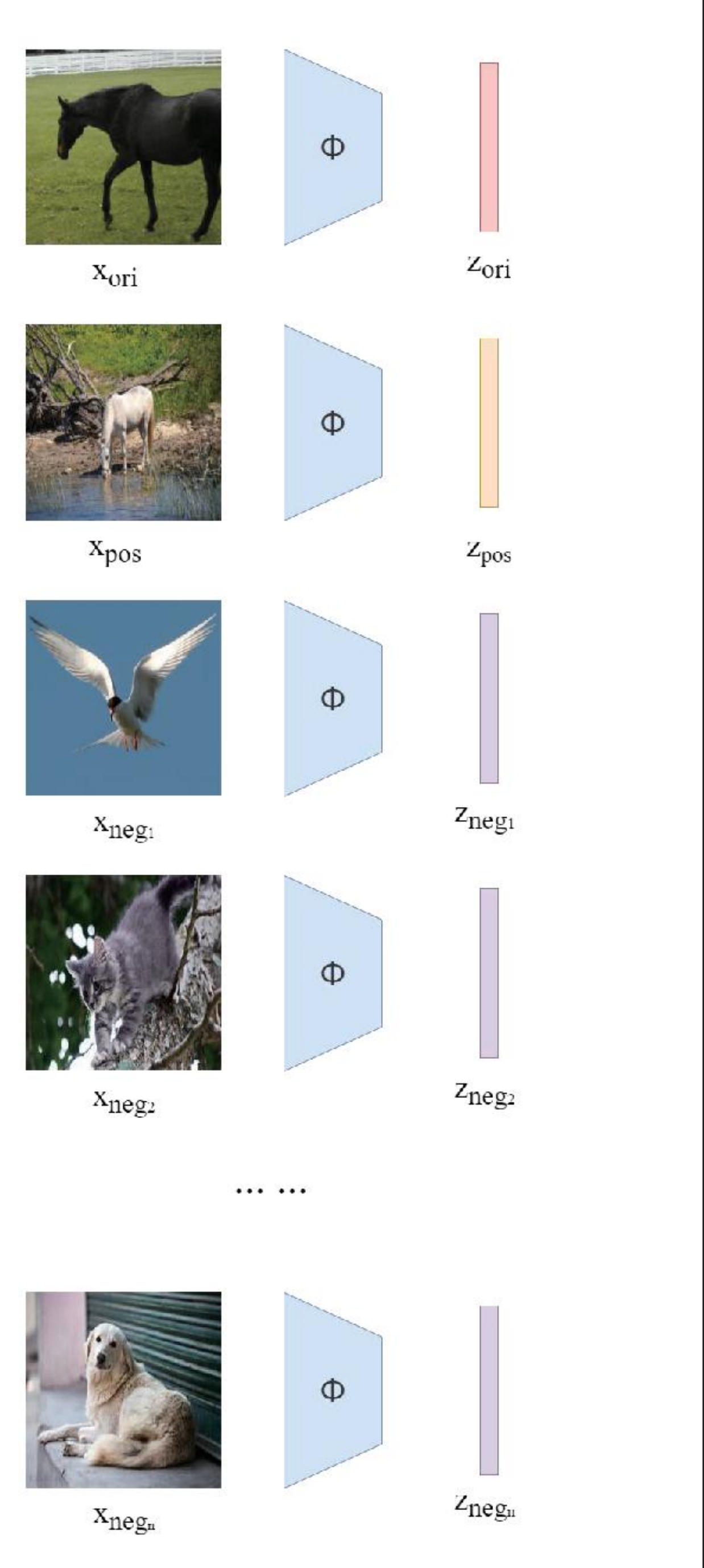}
	\caption{The figure shows the feedforward process of input data.}
\label{loss_function}
\end{figure}

${I } \in {R^{d \times d}}$ in Equation (\ref{cov}) is the identity matrix, and we use $Z \in {R^{n \times d}}$, which is composed of a batch of $z$ to calculate ${\ell _{{\rm{cov}}}}\left( {Z} \right)$. We adopt ancestral sampling approach in each sample of $X_{ori}$ to form $X_{pos}$ and $X_{neg^m}$, and calculate ${\ell _{{\rm{cov}}}}\left( Z \right)$, $ {\ell _{{\rm{cov}}}}\left( {{Z_{pos}}} \right)$ and ${\ell _{{\rm{cov}}}}\left( {{Z_{neg}}} \right)$ separately. We use Figure \ref{loss_function} to describe the feedforward process of input data in more detail.

At the end of this section, we attach a simple comparative experiment to demonstrate the performance of using the first paradigm alone. The experimental data of this section directly adopts the setting in the first experiment, that is, 8 source domains and 2 target domains. The pseudo-code for the comparative learning is shown in {Algorithm 1}. Our model is {not initialized with any pre-trained model}.

\begin{algorithm}[ht]
		\label{Algorithm 1}
		\caption{the first stage of EiHi net}
		\begin{algorithmic}[1]
			\State\textbf{Input:} Training set $D$, number of samples $n$
			\State\qquad number of epoch $epoch$
			\State\qquad feature capture network:${f_\theta }\left( \cdot \right)$
            \State\qquad $M$ is the amount of negative samples in each minimal learning element.
            \State \textbf{Output} Trained ${f_\theta }\left( \cdot \right)$

			\For{$i$ in $(1,2,3,...,epoch)$}
				\State Take \textbf{samples} from $D$:
				\State $X_{ori}=[{x_{ori}}_{_1},{x_{ori}}_{_2},...{x_{ori}}_{_n}];$
				\State According to the category of sample in $X_{ori}$, randomly select a \textbf{positive sample} with $m$ \textbf{negative samples} for each original sample:
				\State $X_{pos}=[{x_{pos}}_{_1},{x_{pos}}_{_2},...{x_{pos}}_{_n}];$
				\State $X_{neg^m}=[{x_{neg^m}}_{_1},{x_{neg^m}}_{_2},...{x_{neg^m}}_{_n}];$
				\State \textbf{feed forward:}
				\State ${Z_{ori}} = {f_\theta }\left( {{X_{ori}}} \right)$;
				\State ${Z_{pos}} = {f_\theta }\left( {{X_{pos}}} \right)$;
				\For{$m$ $in (1, 2, ..., M)$}
					\State \textbf{feed forward:}
					\State ${Z_{neg^m}} = {f_\theta }\left( {{X_{neg^m}}} \right)$;
				\EndFor
				\State \textbf{calculate:}
				\State $\ell = \ell\left( {{Z_{ori}},{Z_{pos}},{Z_{ne{g^1}}},{Z_{ne{g^2}}},...,{Z_{ne{g^M}}}} \right)$ by Equation (\ref{loss})
				\State $\theta \leftarrow \theta - \alpha \cdot \frac{{\nabla \ell }}{{\nabla \theta }}$		
			\EndFor
		\end{algorithmic}
	\end{algorithm}

We train the feature capture network and fully connected nonlinear discriminator separately, that is, the discriminator is trained with the fixed parameters of the feature capture network. In the experiment, 1 positive sample and 9 negative samples were selected to form a minimal learning element with the original sample and we set the $temper$ coefficient to 0.01. The model parameters with the highest accuracy during training are kept for the test. We run each test five times and keep the average results, as shown in Table \ref{vehicle feature extraction}. One should note that {only category labels} are available in our environment. This is different from DecAug and its related work.

\begin{table}[ht]
\centering
\caption{EiHi with first paradigm, experiment on NICO 8:2}
  \begin{threeparttable}
\begin{tabular}{cccccc}
\hline\hline
dataset&backbone&\makecell[c]{without first \\paradigm(\%)}&\makecell[c]{with first\\ paradigm(\%)}\\
\hline

\multirow{4}*{NICO-Animal}&ResNet18  	  &28.56$\pm$0.30 &\textbf{36.79$\pm$0.05}   \\

~&ResNet50	  &\textbf{39.71$\pm$0.30}  &36.43$\pm$0.07          \\

~&Vgg16       	&48.66$\pm$0.20  &\textbf{63.38$\pm$0.04}     \\

~&Vgg19       	&46.11$\pm$0.09  &\textbf{61.45$\pm$0.06}      \\ 

\hline

\multirow{4}*{NICO-Vehicle}&ResNet18  	   &42.29$\pm$0.08 &\textbf{49.89$\pm$0.09}        \\

~&ResNet50	  &\textbf{55.16$\pm$0.35}  &47.82$\pm$0.05          \\

~&Vgg16       	&58.24$\pm$0.40  &\textbf{68.09$\pm$0.14}      \\

~&Vgg19       	&57.60$\pm$0.80  &\textbf{70.42$\pm$0.06}      \\ 

\hline\hline
\end{tabular}
\label{vehicle feature extraction}
 \begin{tablenotes}
        \footnotesize
        \item 8 source domains for training and 2 target domains for testing
      \end{tablenotes}
  \end{threeparttable}
\end{table}

\subsection{Spurious causal elimination}
\begin{figure}[ht]\centering
	\includegraphics[width=8cm]{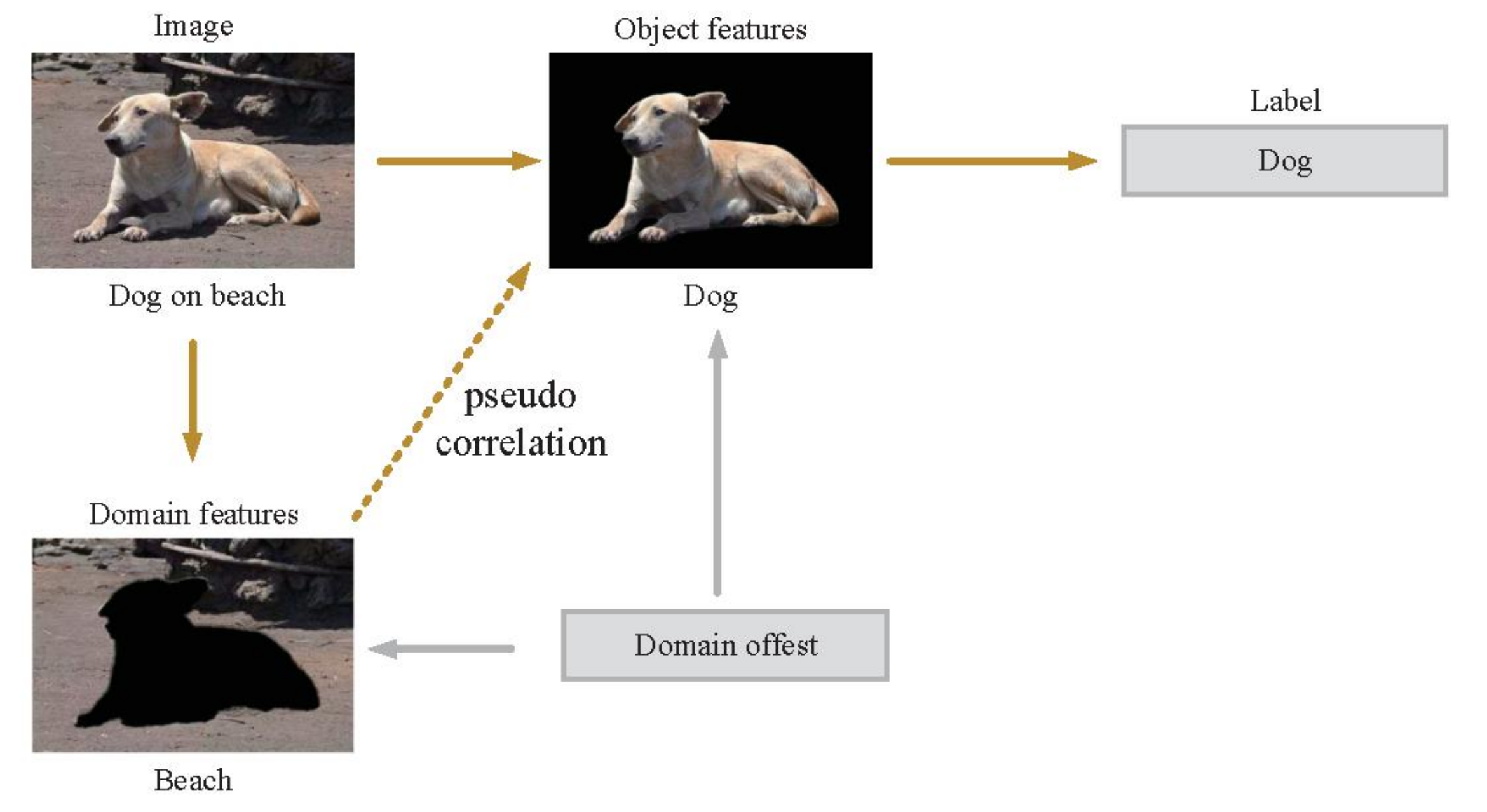}
	\caption{The domain offset factor leads to the generation of wrong causal chain.}
\label{Causal_chain}
\end{figure}

Traditional CNN determines the category according to the features captured by the convolutional layers: it links features with labels in terms of the statistical relevance level. This process leads to the potential formation of false causality \cite{ref30,ref31,ref-1}, because of the skewed domain distribution which leads to false relevance between domain features and labels, producing a wrong causal chain (the causal chain is shown in Figure \ref{Causal_chain}). Eliminating false causality and making domain information {no longer participate} in the decision-making process has become the primary task. Therefore, we use the strategy of {human-in-the-loop} to provide some hints for the deep learning model.

We select very few samples ($e.g.$, 10, the volume is the same as the number of categories) to help eliminate the participation of domain features in the process of the sample category identification. We remove domain information from the selected samples $ X_{sel}$, only retain object information, and bind the samples before and after deleting domain information as guidance sample pair $\left\{ X_{sel},X_{obj}\right\}$(where $X_{sel} = \left\{ {{x_{se{l_k}}}} \right\}_{k = 1}^K$, $X_{obj} = \left\{ {{x_{ob{j_k}}}} \right\}_{k = 1}^K$, ${{x_{se{l_k}}}},{{x_{ob{j_k}}}} \in {R^{c \times h \times w}}$, $K$ is the number of selected samples, ${{x_{ob{j_k}}}}$ is the ${{x_{se{l_k}}}}$ without background information).

To reduce the participation of background information in image category identification, it is desired that the image representation is dimensionally disentangled to some extent. In our work, the covariance term in loss guarantees that each dimension is uncorrelated (since independence is far more difficult to achieve). Specifically, we input the guidance samples $\left\{ X_{sel}, X_{obj}\right\}$ into the feature capture network to observe how the sample features change due to the elimination of background information. The dimensions of features whose values change significantly need to be eliminated.

The process is performed sample pair by sample pair. Denote the feature representation of guidance sample pairs as $Z_{sel}= \left\{ {{z_{se{l_k}}}} \right\}_{k = 1}^K$, $Z_{obj}=\left\{ {{z_{ob{j_k}}}} \right\}_{k = 1}^K$ (${z_{se{l_k}}}, {z_{ob{j_k}}} \in {R^d}$). For each sample pair, we calculate the changing magnitude of feature representations, as shown in Equation \ref{changing_magnitude}:

\begin{align}\label{changing_magnitude}
p_{k}=\frac{{|{z_{se{l_k}}} - {z_{ob{j_k}}}|}}{{||{z_{se{l_k}}}||}}\times 100\%
\end{align}

We then find out dimensions of $p_{k}$ which contains top 90\% of the value (changing magnitude), then index these dimensions in $p_{k}$ as 0, while index the rest dimensions in $p_{k}$ as 1. By doing so, we find the top 90\% dimensions in the original feature representation which may be sensitive to the background. $P = \left\{ {p_{k}} \right\}_{k = 1}^K$ is obtained once we finish processing all the guidance sample pair. Then we define $I_{guid}$ to make an overall decision in terms of all the guidance sample pairs, as shown in Equation \ref{index_elimination}

\begin{align}\label{index_elimination}
I_{guid} = \sum\nolimits_{k = 1}^K {{p_k}}
\end{align}

\begin{figure}[ht]\centering
	\includegraphics[width=8cm]{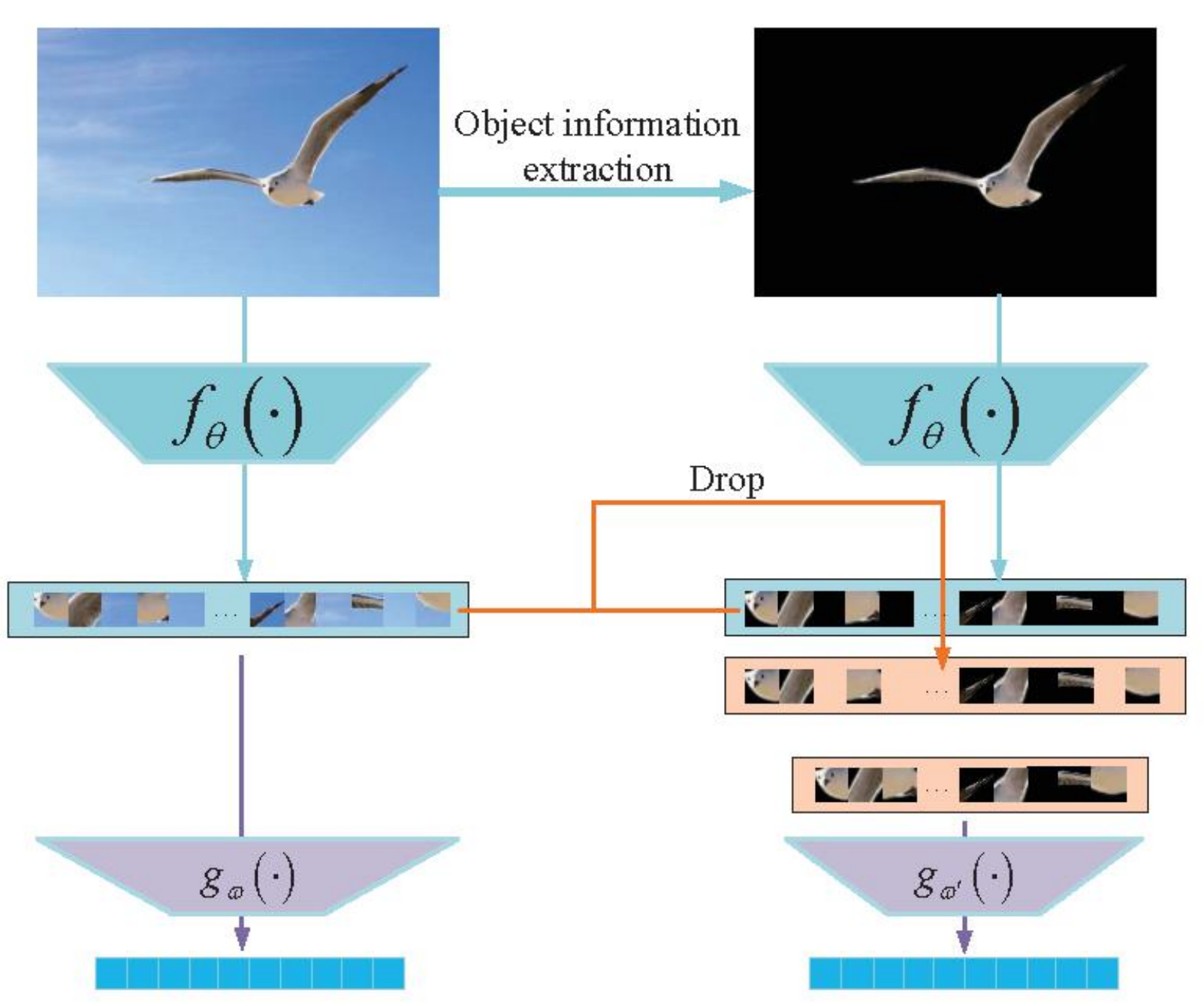}
	\caption{The figure shows the process of eliminating the domain information dimension.}
\label{drop_fig}
\end{figure}

$I_{guid} \in R^{d}$ is an indicator, and dimension with 0 value in $I_{guid}$ indicates that all the guidance sample pairs agree that this dimension is sensitive to the background. Finally, we keep the feature capture network fixed and retrain the nonlinear discriminator, the feature dimensions which are sensitive to the background will be opted out during training, with the information from $I_{guid}$. The whole process can be seen in Figure \ref{drop_fig}. Note the the `opt out' operation of dimension means the elimination of dimension, not setting the value of that dimension to 0.

We designed experiments to show the performance of the proposed paradigm in this section. We selected {10} samples in NICO-animal and {9} samples in NICO-vehicle as guidance samples, the amount of chosen samples are consistent with the number of categories. The experiment setting is consistent with the previous experiment. The results are shown in Table \ref{Remove vehicle spurious correlation features}.

\begin{table}[ht]
\centering
\caption{Context feature droping experiment on NICO 8:2}
\begin{threeparttable}
\begin{tabular}{cccccc}
\hline\hline
dataset&backbone&\makecell[c]{without first \\paradigm(\%)}&\makecell[c]{only second\\ paradigm(\%)}&\\
\hline
\multirow{4}*{NICO-Animal}&ResNet18  	  &28.56$\pm$0.30 &\textbf{30.52$\pm$0.04}       \\

~&ResNet50	  &39.71$\pm$0.30  &\textbf{40.50$\pm$0.29}         \\

~&Vgg16       	&48.66$\pm$0.20  &\textbf{57.53$\pm$0.44}    \\

~&Vgg19       	&46.11$\pm$0.09  &\textbf{52.90$\pm$0.20}      \\

\hline

\multirow{4}*{NICO-Vehicle}&ResNet18  	   &42.29$\pm$0.08 &\textbf{49.27$\pm$0.27}       \\

~&ResNet50	  &\textbf{55.16$\pm$0.35}  &52.05$\pm$0.30          \\

~&Vgg16       	&58.24$\pm$0.40  &\textbf{63.52$\pm$0.14}     \\

~&Vgg19       	&57.60$\pm$0.80  &\textbf{64.44$\pm$0.06}      \\

\hline\hline
\end{tabular}
\label{Remove vehicle spurious correlation features}
 \begin{tablenotes}
        \footnotesize
        \item 8 source domains for training and 2 target domains for testing
      \end{tablenotes}
  \end{threeparttable}
\end{table}

\subsection{EiHi Net}
We combine the feature capture process and spurious causal elimination process into our EiHi net. Different instantiations of Vgg and ResNet are used as backbones in the feature capture network. In our experiment, only object labels are available and we use limited human efforts in eliminating background information for the second paradigm of EiHi. We list the test results of EiHi on these backbones without pre-training on NICO dataset in table \ref {EiHi_nico1}. We have also made a comparison of various algorithms. For comparison, we directly cite the average results of the StableNet on two NICO datasets recorded in \cite{ref13}. The results of the comparison are shown in Table \ref{horizontal comparison}. The SCM's number of weight vector $t$ is set to $10$, which is the same as the NICO's domains.
\begin{table}[ht]
\centering
\caption{EiHi net on NICO 8:2}
\begin{threeparttable}
\begin{tabular}{cccccc}
\hline\hline
dataset &backbone&\makecell[c]{only\\ backbone(\%)}&\makecell[c]{EiHi net(\%)}&\\
\hline

\multirow{4}*{NICO-Animal}&ResNet18  	  &28.56$\pm$0.30   &37.42$\pm$0.06      \\

~&ResNet50	  &39.71$\pm$0.30      &37.21$\pm$0.09         \\

~&Vgg16       	&48.66$\pm$0.20    &\textbf{64.41$\pm$0.05}       \\

~&Vgg19       	&46.11$\pm$0.09    &62.80$\pm$0.04      \\ 

\hline
\multirow{4}*{NICO-Vehicle}&ResNet18  	   &42.29$\pm$0.08  &51.89$\pm$0.04        \\

~&ResNet50	  &55.16$\pm$0.35    &48.87$\pm$0.10           \\

~&Vgg16       	&58.24$\pm$0.40  &69.71$\pm$0.05      \\

~&Vgg19       	&57.60$\pm$0.80  &\textbf{71.34$\pm$0.04}      \\ 

\hline\hline
\end{tabular}
\label{EiHi_nico1}
 \begin{tablenotes}
        \footnotesize
        \item 8 source domains for training and 2 target domains for testing
      \end{tablenotes}
  \end{threeparttable}

\end{table}

\begin{table}[ht]
\centering
\caption{various net on NICO 8:2}
\begin{threeparttable}
\begin{tabular}{cccc}
\hline\hline
data set &model&\makecell[c]{average accuracy}&\\
\hline
\multirow{7}*{\makecell[c]{NICO-Animal \\and\\ NICO-Vehicle}}&JiGen    &54.72   \\

~&M-ADA	  &40.78        \\

~&DG-MMLD     	&47.18   \\

~&RSC      	&57.59      \\ 

~&\makecell[c]{ResNet-18(pre-training)} &51.71 \\

~&StableNet   &59.76 \\

~&SCM   &{35.75} \\

~&EiHi(ours)   &\textbf{67.88} \\

\hline\hline
\end{tabular}
\label{horizontal comparison}
 \begin{tablenotes}
        \footnotesize
        \item 8 source domains for training and 2 target domains for testing
      \end{tablenotes}
  \end{threeparttable}
\end{table}

\begin{figure}[ht]\centering
	\includegraphics[width=8cm]{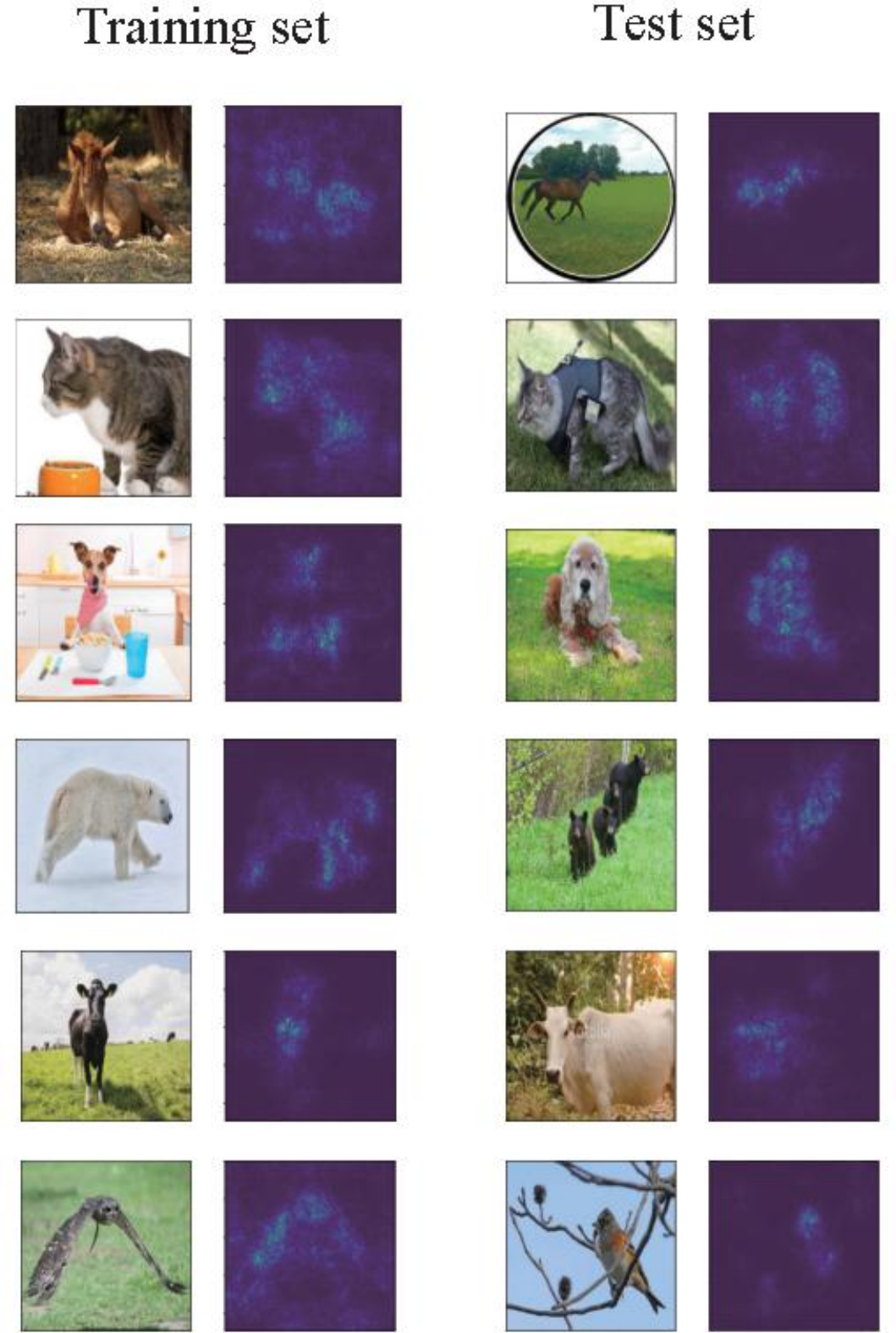}
	\caption{The figure shows the saliency map of EiHi net for each NICO-Animal sample.}
\label{heat_map}
\end{figure}

\begin{figure}[ht]\centering
	\includegraphics[width=8cm]{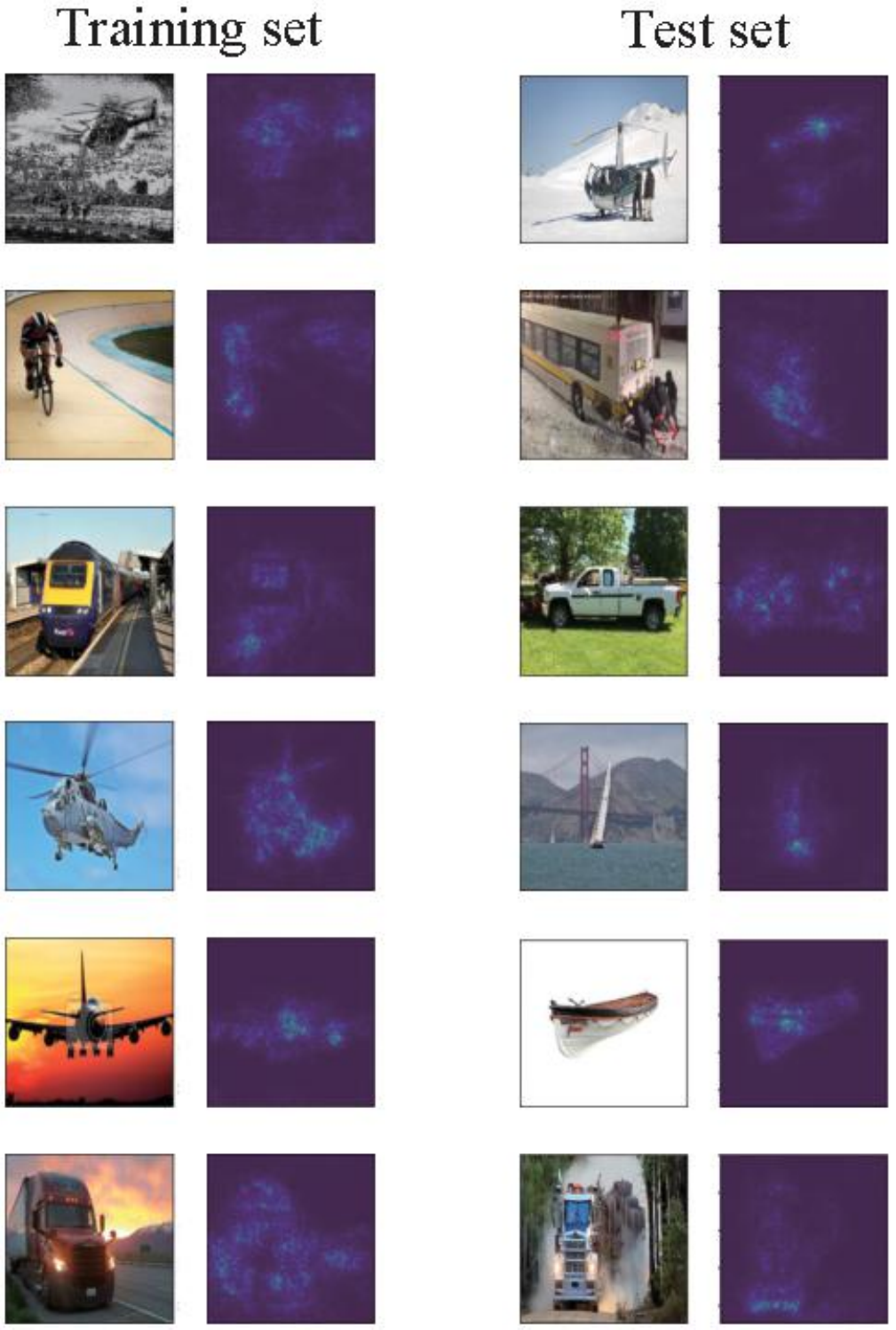}
	\caption{The figure shows the saliency map of EiHi net for each NICO-Vehicle sample.}
\label{heat_map_vehicle}
\end{figure}

We selected some samples in the training set and test set, and used EiHi net to identify them. Then we drew the saliency map \cite{saliency} of EiHi net for each sample. Some examples of saliency maps are shown in Figures \ref{heat_map} and \ref{heat_map_vehicle}.

\section{Challenge}
The first challenge is the greater Diversity Shift. We try an enhanced Diversity Shift by training 7 domains and testing 3 domains in NICO. The experimental results are recorded in Table \ref{strengthen Diversity Shift}.

\begin{table}[ht]
\centering
\caption{EiHi net on NICO 7:3}
\begin{threeparttable}
\begin{tabular}{ccccc}
\hline\hline
\makecell[c]{data set}&model&\makecell[c]{Vgg16(\%)}&\makecell[c]{EiHi net(\%)}\\
\hline

\makecell[c]{NICO-Animal}&\makecell[c]{EiHi net\\(Vgg16) }      	&46.13$\pm$0.85  &\textbf{62.26$\pm$0.05}\\
\makecell[c]{NICO-Vehicle}&\makecell[c]{EiHi net\\(Vgg16)}      	&55.62$\pm$1.60  &\textbf{62.72$\pm$0.08}\\

\hline\hline
\end{tabular}
\label{strengthen Diversity Shift}
 \begin{tablenotes}
        \footnotesize
        \item Re-select 7 source domains for training and 3 target domains for testing
      \end{tablenotes}
  \end{threeparttable}
\end{table}

This paper tries to strengthen the correlation shift to cope with another challenge. We select one of the 8 source domains as the primary domain and set other domains as the secondary domain for each class. We set the ratio of the number of primary domain samples to the number of secondary domain samples to 5:1. This ratio is based on the {Adversarial} setting in \cite{ref13}. The original intention of the Adversarial setting is the same as the purpose of enhancing the correlation shift. The results can be seen in Table \ref{strengthen correlation shift}. Other model results appearing in Table \ref{strengthen correlation shift} are from reference \cite{ref12}.

\begin{table}[ht]
\centering
\caption{EiHi on NICO 8:2 with domain shift 5:1:1:1:1:1:1:1}
\begin{threeparttable}
\begin{tabular}{ccccc}
\hline\hline
\makecell[c]{dataset}&model&\makecell[c]{accuracy}\\
\hline
\multirow{6}*{\makecell[c]{NICO-Animal}}&\makecell[c]{CNN}       	&37.17  \\
~&\makecell[c]{CNN+BN}       	&38.70   \\
~&\makecell[c]{CNBB}       	&\underline{39.06}   \\
~&\makecell[c]{pure backbone\\(vgg16)}    	&37.83   \\
~&~&~&~&~\\
~&\makecell[c]{EiHi net\\(vgg16)}   &\textbf{57.68$\pm$0.05}  \\
\hline
\multirow{6}*{\makecell[c]{NICO-Vehicle}}&\makecell[c]{CNN}       	&40.61 \\
~&\makecell[c]{CNN+BN}       	&\underline{41.98}   \\
~&\makecell[c]{CNBB}       	&41.41   \\
~&\makecell[c]{pure backbone\\(vgg16)}       	&41.85   \\
~&~&~&~&~\\
~&\makecell[c]{EiHi net\\(vgg16)}       	&\textbf{57.86$\pm$0.09 }  \\ 

\hline\hline
\end{tabular}
\label{strengthen correlation shift}
 \begin{tablenotes}
        \footnotesize
        \item Other model results appearing in Table \ref{strengthen correlation shift} are from reference \cite{ref12}.
	\item The data volume of the secondary domain in the source domain is reduced by $1/5$ of the primary domain.
      \end{tablenotes}
  \end{threeparttable}

\end{table}

\section{Ablation Study}
In our approach. Our main loss functions are $infoNCE$ and the covariance term. First, we evaluate the effect of EiHi without the covariance term. Second, we report the results of $VICReg$ on NICO (run by ourselves) to show that it is not applicable to small data presenting O.o.D problems.

\begin{table}[ht]
\centering
\caption{Ablation study on NICO 8:2}
\begin{threeparttable}
\begin{tabular}{cccccc}
\hline\hline
dataset &baceline&\makecell[c]{result(\%)}\\
\hline

\multirow{4}*{NICO-Animal}&\makecell[c]{EiHi'stage one\\(without Covariance term) }       	&46.16$\pm$0.04           \\
~&\\

~&VICReg\tnote{1}  	  &25.40$\pm$0.05\\
~&\\

~&EiHi'stage one       	&\textbf{63.38$\pm$0.04 }      \\ 

\hline
\multirow{4}*{NICO-Vehicle}&\makecell[c]{EiHi'stage one\\(without Covariance term)  }     	&60.47$\pm$0.03        \\

~&\\
~&VICReg\tnote{1} 	   &30.50$\pm$0.03       \\
~&\\

~&EiHi'stage one      	&\textbf{68.09$\pm$0.14}      \\ 

\hline\hline
\end{tabular}
\label{EiHi Abl}
 \begin{tablenotes}
        \footnotesize
        \item{1} When training VICReg, the batchsize is set to 990, and no sample pair is set.

      \end{tablenotes}
  \end{threeparttable}
\end{table}

\section{Conclusion}
In this paper, we propose the EiHi net, a paradigm that can be blessed on any backbone, to deal with the background O.o.D. generalization tasks. With limited human efforts, when using our paradigm to train the backbone, there is no need for additional domain labels and pre-training parameter loading. With category labels are the only available information, we achieve competitive results on the most difficult and representative two-dimensional O.o.D. dataset NICO. We solve the problem of NICO without adding the pre-training model. In addition, avoiding labeling domain information and inputting it into the deep learning model as a supervision signal saves manpower.

\section{Style Information O.o.D.}
PACS\cite{refpacs} as a typical style information O.o.D. database, which records seven kinds of things in four styles (photos, paintings, cartoons, simple strokes). its O.o.D. mode is different from the background O.o.D. problem solved by EiHi. Quality work\cite{ref11}\cite{ref13} based on this database depends on pretraining and style information tags. EiHi replaced the backbone with ResNet18 and tried it on PACS with pre-training. the result of EiHi and the results of methods that do not use style tags are shown in Tabel\ref{pacs}. Our results only shows limited competitiveness to some extent. However, we still believe that excessive dependence on the pre-training model is not the final solution to the O.o.D. problem.
Different from the background O.o.D problem, there is no `wrong feature' in the style O.o.D. data (what they have is some features useful in one situation is not useful in other situations). EiHi is used to suppress wrong features based on sample pairs, which we think is the reason why EiHi is not suitable for style O.O.D. problems.

\begin{table}[ht]
\centering
\caption{Ablation study on PACS}
\begin{tabular}{ccccccc}
\hline\hline
baceline &Art&Cartoon&Sketch&Photo\\
\hline
JiGen &  79.42 &75.25 &71.35 &96.03\\
M-ADA &  64.29 &72.91 &67.21 &88.23\\

DG-MMLD &  81.28 &77.16 &72.29 &96.09       \\

REC    &   83.43 &80.31 &80.85 &95.99\\

ResNet-18  &76.61 &73.60 &76.08 &93.31 \\

Stable Net &     81.74 &79.91 &80.50 &96.53\\

EiHi stage one &80.57 & 75.64& 71.09 &95.57\\

\hline\hline
\end{tabular}
\label{pacs}
\end{table}


\begin{IEEEbiography}[{\includegraphics[width=1in,height=1.25in,clip,keepaspectratio]{pic/Author_Qinglaiwei.eps}}]
{Qinglai Wei} received the B.S. degree in Automation, and the Ph.D. degree in control theory and control engineering, from the Northeastern University, Shenyang, China, in 2002 and 2009,
respectively. From 2009--2011, he was a postdoctoral fellow with The State Key Laboratory of Management and Control for Complex Systems, Institute of Automation, Chinese Academy of Sciences, Beijing, China. He is currently a professor of the institute and the associate director of the laboratory. He has authored six books, and published over 100 international journal papers. His research interests include artificial intelligence, deep learning, adaptive dynamic programming, neural networks, optimization, and their applications.
\end{IEEEbiography}

\begin{IEEEbiography}[{\includegraphics[width=1in,height=1.25in,clip,keepaspectratio]{pic/Author_Beimingyuan.eps}}]
{Beiming Yuan} received the B.S. degree in computer science and technology from Shenyang Ligong University. He is currently pursuing the M.S. degree in Electronic information with the University of Chinese Academy of Sciences. His current research interests include stable learning.
\end{IEEEbiography}

\begin{IEEEbiography}[{\includegraphics[width=1in,height=1.25in,clip,keepaspectratio]{pic/Author_Dianchengchen.eps}}]
{Diancheng Chen} received the B. Eng. degree in automation engineering from Dalian University of Technology, Dalian, China, in 2017, and the MSc degree in Mechanical and Automation Engineering from The Chinese University of Hong Kong, Hong Kong, China, in 2018. He is currently pursuing the Ph.D. degree in control theory and control engineering with the State Key Laboratory of Management and Control for Complex Systems, Institute of Automation, Chinese Academy of Sciences, Beijing, and the University of Chinese Academy of Sciences, Beijing. His current research interests include abstract reasoning, deep learning.
\end{IEEEbiography}

\end{document}